%% file: main_text.tex
\documentclass[conference]{IEEEtran}
\IEEEoverridecommandlockouts
\usepackage{cite}
\usepackage{epsfig}
\usepackage{amsmath,amssymb,amsfonts}
\usepackage{algorithmic}
\usepackage{graphicx}
\usepackage{textcomp}
\usepackage{pgfplots}
\usepackage{tabularx}
\usepackage{hyperref}

\usetikzlibrary{spy,calc}
\usepackage{booktabs}

\usepackage{xcolor}
\usepackage{tikz}
\def\BibTeX{{\rm B\kern-.05em{\sc i\kern-.025em b}\kern-.08em
    T\kern-.1667em\lower.7ex\hbox{E}\kern-.125emX}}
\begin{document}

\title{MetaFed: Advancing Privacy, Performance, and Sustainability in Federated Metaverse Systems\
}





\author{
\begin{tabular}{ccc}
{Muhammet Anil Yagiz} & {Zeynep Sude Cengiz} & {Polat Goktas} \\[0.3em]
\textit{Department of Computer Engineering} & \textit{Department of Computer Engineering} & \textit{Faculty of Engineering and} \\[0.3em]
\textit{Kırıkkale University} & \textit{Kırıkkale University} & \textit{Natural Sciences, Sabancı University} \\[0.3em]
Kırıkkale, Türkiye & Kırıkkale, Türkiye & Istanbul, Türkiye \\[0.3em]
 & & 0000-0001-7183-6890
\end{tabular}
}

\maketitle

\begin{abstract}
The rapid expansion of immersive Metaverse applications introduces complex challenges at the intersection of performance, privacy, and environmental sustainability. Centralized architectures fall short in addressing these demands, often resulting in elevated energy consumption, latency, and privacy concerns. This paper proposes \textit{MetaFed}, a decentralized federated learning (FL) framework that enables sustainable and intelligent resource orchestration for Metaverse environments. MetaFed integrates (i) multi-agent reinforcement learning for dynamic client selection, (ii) privacy-preserving FL using homomorphic encryption, and (iii) carbon-aware scheduling aligned with renewable energy availability. Evaluations on MNIST and CIFAR-10 using lightweight ResNet architectures demonstrate that MetaFed achieves up to 25\% reduction in carbon emissions compared to conventional approaches, while maintaining high accuracy and minimal communication overhead. These results highlight MetaFed as a scalable solution for building environmentally responsible and privacy-compliant Metaverse infrastructures. The source code and supplementary materials supporting this study are publicly accessible via our GitHub repository.\footnote{Available at: \url{https://github.com/afrilab/MetaFed-FL}}

\end{abstract}

\begin{IEEEkeywords}
Federated Learning, Metaverse, Sustainability, Privacy, Resource Orchestration, Reinforcement Learning
\end{IEEEkeywords}

\section{Introduction}
The Metaverse is emerging as a transformative paradigm, enabling immersive and persistent virtual experiences through an amalgamation of technologies such as extended reality (XR), blockchain, and digital twins \cite{xu2022survey,wang2022metaverse}. These environments rely heavily on real-time rendering, multi-modal interaction, and high-fidelity simulation, often requiring up to 50 times more compute than traditional applications \cite{chen2022resource}. Supporting this scale of computation, especially across millions of users, necessitates robust infrastructures that are not only high-performing but also energy-efficient and privacy-conscious.

Contemporary cloud-centric models suffer from significant drawbacks. First, the centralization of data and resources leads to severe \textit{scalability bottlenecks}, as data centers struggle to manage latency-sensitive and compute-heavy Metaverse workloads \cite{satyanarayanan2017emergence, agrawal2025iot, agrawal2025mohaf}. Second, the aggregation of sensitive user data, including biometric, behavioral, and spatial information, triggers \textit{serious privacy concerns}, especially in jurisdictions governed by strict data protection laws \cite{zhang2021security}. Third, the \textit{environmental impact} of these infrastructures is non-trivial; data centers are already responsible for 3--4\% of global electricity use, with projections showing further escalation driven by artificial intelligence (AI) and XR \cite{strubell2019energy,masanet2020recalibrating}.

The unique demands of Metaverse environments amplify these challenges beyond traditional distributed computing scenarios. Unlike conventional applications, Metaverse platforms require real-time processing of multi-modal sensory data, continuous synchronization of shared virtual states across thousands of concurrent users, and ultra-low latency responses to maintain immersion \cite{wang2022metaverse}. These requirements translate to computational workloads that are not only massive in scale but also highly dynamic and unpredictable, making traditional static resource allocation strategies fundamentally inadequate.

To address these intertwined challenges, we propose \textbf{MetaFed}, a decentralized, intelligent orchestration framework tailored for federated Metaverse computing. Our approach integrates three foundational innovations specifically designed for Metaverse infrastructural needs:
\begin{itemize}
    \item A \textit{multi-agent reinforcement learning (MARL)}-based orchestration engine that dynamically adapts resource allocation based on system states, sustainability metrics, and privacy policies. MARL-based dynamic resource allocation is critically important for maintaining fluid user experiences under the constantly shifting user loads and interaction demands characteristic of Metaverse environments \cite{chen2022resource}.
    \item A \textit{privacy-by-design federated learning (FL) architecture} incorporating homomorphic encryption and differential privacy mechanisms to protect user data and model gradients. In Metaverse environments where sensitive biometric and behavioral data are continuously processed \cite{zhang2021security}, FL enhanced with homomorphic encryption and differential privacy serves as an indispensable privacy shield.
    \item \textit{Carbon-aware scheduling strategies} that leverage real-time energy grid data to prioritize low-emission computation, minimizing environmental footprint without compromising performance. Given the massive energy footprint potential of Metaverse infrastructure \cite{strubell2019energy}, carbon-aware scheduling becomes a necessity rather than an option for ensuring environmental sustainability of this transformative technology.
\end{itemize}

Thus, \textit{MetaFed} aims to provide a robust response to the unique needs of Metaverse environments, enabling scalable, private, and sustainable orchestration across heterogeneous edge and cloud infrastructures.

\section{Related Work}
\subsection{Federated Learning and Distributed AI}
FL has become the de facto framework for distributed machine learning (ML) without raw data centralization \cite{mcmahan2017communication}. Techniques such as FedProx \cite{li2020federated}, SCAFFOLD \cite{karimireddy2020scaffold}, and FedNova \cite{wang2020tackling} have attempted to address statistical heterogeneity and slow convergence. More recent works explore personalization \cite{tan2022towards} and byzantine robustness \cite{blanchard2017machine}. However, these approaches primarily focus on training efficiency and client reliability, without considering holistic orchestration or energy implications.

\subsection{Resource Management in Edge Computing}
Edge computing facilitates latency-sensitive services by distributing compute close to data sources \cite{shi2016edge,mao2017survey}. Resource orchestration strategies include dynamic task offloading \cite{chen2019edge}, bandwidth-aware scheduling \cite{wang2017bandwidth}, and load balancing \cite{li2018learning}. Reinforcement learning (RL)-based techniques \cite{li2021deep} and game-theoretic models \cite{zeng2016joint} have shown promise in adaptive orchestration. However, they often optimize single objectives, such as latency or throughput, and rarely incorporate privacy or sustainability constraints.

\subsection{Sustainable and Green AI}
The energy cost of large-scale AI has led to the rise of Green AI, advocating for transparency in reporting energy use and promoting eco-efficient design \cite{strubell2019energy, lacoste2019quantifying}. Tools and methods for carbon-aware job scheduling \cite{li2011carbon}, renewable integration \cite{goiri2013parasol}, and geographical load balancing \cite{qureshi2009cutting} have been proposed. Recent studies explore energy-efficient FL \cite{zeng2021energy} and sustainable ML pipelines \cite{bannour2021evaluating}, but lack integration with real-time orchestration or privacy protection.

\subsection{Privacy-Enhancing Technologies in Federated Systems}
Maintaining user privacy in decentralized settings is critical. Techniques like differential privacy \cite{abadi2016deep}, homomorphic encryption \cite{gentry2009fully}, and secure aggregation protocols \cite{bonawitz2017practical} offer varying trade-offs between computational overhead and privacy guarantees. While federated systems have begun integrating these methods, combining them with dynamic orchestration and environmental constraints remains underexplored.

\subsection{Computing for Metaverse Infrastructures}
Recent surveys highlight the gaps in computing paradigms for the Metaverse \cite{xu2022survey,lee2024all}. While studies address immersive networking \cite{yukun2024computing}, mobile edge integration \cite{wang2022survey}, and rendering acceleration \cite{wang2022metaverse}, few offer holistic solutions that jointly optimize performance, privacy, and sustainability.  Existing Metaverse infrastructure proposals typically focus on single-objective optimization, either maximizing rendering performance, minimizing latency, or reducing costs, without considering the complex interplay between computational efficiency, user privacy, and environmental impact \cite{baidya2024comprehensive}. This fragmented approach fails to address the fundamental challenge that Metaverse platforms must simultaneously deliver exceptional user experiences while respecting privacy regulations and sustainability constraints.

Therefore, MetaFed distinguishes itself by providing comprehensive, multi-objective optimization that treats performance, privacy, and sustainability as equally critical design constraints rather than optional considerations. Our proposed framework addresses this key research gap by offering the first principled approach to holistic Metaverse infrastructure orchestration.

\vspace{0.5em}
\noindent In summary, the current landscape in the literature offers valuable contributions in FL, privacy, and sustainable computing, but fails to comprehensively address the triadic challenge of intelligent, private, and green resource orchestration in federated Metaverse systems. \textit{MetaFed} fills this gap with a principled, multi-objective orchestration architecture.

\section{Methodology}
MetaFed operationalizes a three-pronged orchestration pipeline tailored to federated Metaverse environments. It fuses intelligent resource selection, privacy-preserving model updates, and green computing practices into a decentralized architecture. This section details the computational framework, including formal state definitions, action models, and scheduling strategies.

\subsection{System Architecture Overview}
The system comprises the following core components:
\begin{itemize}
    \item \textbf{Resource Providers ($\mathcal{R}$)}: Edge and cloud nodes that contribute computational capacity, energy efficiency metrics, and location-based carbon intensity data.
    \item \textbf{Orchestration Agents ($\mathcal{A}$)}: Multi-agent reinforcement learners that perform decentralized decision-making for task orchestration.
    \item \textbf{Federated Registry ($\mathcal{F}$)}: A blockchain-based distributed hash table that stores metadata for resource discovery and identity verification.
\end{itemize}

Each resource provider $r_i \in \mathcal{R}$ is represented as:
\begin{equation}
    r_i = \langle \mathcal{C}_i, \mathcal{N}_i, \mathcal{E}_i, \mathcal{L}_i \rangle
\end{equation}
where $\mathcal{C}_i$ is normalized computational capability, $\mathcal{N}_i$ is network bandwidth, $\mathcal{E}_i$ reflects energy efficiency, and $\mathcal{L}_i$ provides geolocation for emission modeling.

\subsection{AI-Driven Orchestration Engine}
The system state at time $t$ is modeled as:
\begin{equation}
    s_t = \langle C_t, A_t, \mathcal{H}_t \rangle
\end{equation}
where $C_t$ is the carbon intensity class (\textit{low, medium, high}), \linebreak $A_t$ is the trend of model accuracy (\textit{up, down}), and $\mathcal{H}_t$ includes metrics like convergence velocity and utilization history.

Each agent chooses from an action space \linebreak $\mathcal{A}_t = \{a_1, a_2, \ldots, a_k\}$ with policy $\pi(a|s)$ defined as:
\begin{equation}
\pi(a|s) = \begin{cases}
\arg\max_a Q(s,a) & \text{w.p. } 1-\epsilon \\
\text{Uniform}(\mathcal{A}) & \text{w.p. } \epsilon
\end{cases}
\end{equation}
where $\epsilon_{t+1} = \max(\epsilon_{\min}, \epsilon_t \cdot \gamma)$, with $\epsilon_{\min} = 0.01$ and \linebreak $\gamma = 0.98$.

The reward function balances accuracy ($\Delta A_t$), efficiency ($\Delta E_t$), and emissions ($C_{\text{CO}_2,t}$):
\begin{equation}
R_t = \alpha \cdot \Delta A_t + \beta \cdot \Delta E_t - \gamma \cdot C_{\text{CO}_2,t}
\end{equation}
with $\alpha = 15$, $\beta = 5$, and $\gamma = 1$. To introduce environmental bias, a green-aware correction is applied:
\begin{equation}
Q'(s,a) = Q(s,a) - \lambda \cdot \frac{\mathcal{C}_i - 1.0}{\sigma_C} \cdot \frac{I_{\text{carbon}}}{I_{\text{avg}}}
\end{equation}
where $\lambda = 0.05$, $\sigma_C$ is the standard deviation of compute capability, and $I_{\text{avg}} = 150$ gCO$_2$/kWh.

\subsection{Privacy-Preserving Federated Optimization}
MetaFed integrates adaptive variants of FL, including:

\paragraph{Enhanced FedAvg} Aggregation is performed via:
\begin{equation}
\mathbf{w}_{\text{global}}^{t+1} = \sum_{i \in \mathcal{S}_t} \frac{n_i}{\sum_{j \in \mathcal{S}_t} n_j} \mathbf{w}_i^{t+1}
\end{equation}
with $\mathcal{S}_t$ as the selected client set, and $n_i$ as local data size.

\paragraph{FedProx} To address heterogeneity, we use:
\begin{equation}
\min_{\mathbf{w}} F_i(\mathbf{w}) + \frac{\mu_i}{2}\|\mathbf{w} - \mathbf{w}^t\|^2
\end{equation}
where $\mu_i = \mu_{\text{base}} \cdot (2.0 - \mathcal{C}_i)$.

\paragraph{Homomorphic Encryption:} Clients encrypt gradients using additive homomorphic schemes. Together with noise perturbation, we ensure $(\epsilon=1.2, \delta=10^{-5})$-differential privacy.

\subsection{Carbon-Aware Scheduling}
Grid carbon intensity $I(t)$ is modeled as:
\begin{equation}
I(t) = I_{\text{base}} + A \cdot \sin\left(\frac{2\pi t}{T} + \phi\right) + \epsilon(t)
\end{equation}
with $I_{\text{base}}=150$, $A=70$, $T=24$ hours, and noise $\epsilon(t) \sim \mathcal{N}(0, \sigma^2)$.

The scheduling priority function is:
\begin{equation}
\text{Priority}(i,t) = \frac{Q(s_t, i)}{\max(1, I(t)/I_{\text{threshold}})}
\end{equation}
where $I_{\text{threshold}} = 100$ gCO$_2$/kWh. 
This design enables model aggregation to favor nodes powered by greener energy, thereby reducing lifecycle emissions of FL.

\section{Component-Wise Performance Evaluation}
To assess the individual and synergistic contributions of \textit{MetaFed}’s core components, we design a structured ablation study across multiple system configurations and datasets. The goal is to isolate the effects of intelligent orchestration, green computing, and privacy-preserving FL.

\subsection{Evaluation Configurations}
We define the following experimental variants:

\begin{itemize}
    \item \textbf{MetaFed (RL + Green + ResNet Tiny (RT))}: The complete system combining AI-driven orchestration with carbon-aware scheduling, used as the baseline reference configuration.
    \item \textbf{MetaFed (RL + RT)}: Incorporates only the RL-based orchestration engine without carbon-intensity-aware scheduling.
    \item \textbf{MetaFed (Green + RT)}: Enables only carbon-aware scheduling with random orchestration policy.
    \item \textbf{Baseline FL Models}: Consists of standard FL approaches like FedAvg, FedProx, and FedAdam, without incorporating intelligent orchestration or sustainability-aware mechanisms.
    
\end{itemize}

All configurations utilize a lightweight RT architecture with 4.8 million parameters to ensure fair comparison across methods. The experiments were conducted with heterogeneous clients, simulated using Dirichlet-distributed data partitions ($\alpha = 0.5$) to emulate the non-IID characteristics commonly observed in real-world FL scenarios. The computational setup employed NVIDIA Tesla P100 GPUs with 16GB HBM2 memory, ensuring the high-performance environment required to rigorously evaluate MetaFed’s orchestration efficiency under realistic Metaverse workload conditions.

\subsection{Performance Metrics}
We evaluate each configuration on three axes:

\begin{itemize}
    \item \textbf{Model Accuracy}: Final test accuracy measured on centralized validation data.
    \item \textbf{Carbon Emissions}: Total estimated CO$_2$ emissions based on modeled energy consumption and real-time carbon intensity data.
    \item \textbf{Communication Overhead}: Cumulative model size exchanged during all communication rounds, reflecting network load.
\end{itemize}

\subsection{Datasets and Training Protocol}
We utilize two benchmark image classification datasets:

\begin{itemize}
    \item \textbf{MNIST:} 60,000 training and 10,000 test images (28×28) of handwritten digits.
    \item \textbf{CIFAR-10:} 50,000 training and 10,000 test images (32×32) across 10 object classes.
\end{itemize}

Each federated round selects 10 out of 50 clients (20\% participation), with each client performing 5 local epochs using a local batch size of 32. Training continues for 100 communication rounds.

\subsection{Experimental Design Rationale}
Although MNIST and CIFAR-10 are conventional benchmark datasets, their selection in this study is a deliberate methodological choice aimed at isolating and evaluating the orchestration framework's performance under controlled and interpretable conditions. These datasets offer a well-understood and stable testing ground, allowing us to focus specifically on the orchestration trade-offs among performance, privacy, and sustainability, without the confounding effects introduced by the complexity or domain-specific nuances of more intricate datasets. To realistically simulate the challenges of FL in Metaverse environments, we employed a Dirichlet distribution with a concentration parameter of \( \alpha = 0.5 \), generating heterogeneous, non-IID data partitions across clients. This setup closely mirrors the variability of user-generated data in practical Metaverse deployments, where individual users or edge devices contribute uneven and highly personalized data patterns.

In addition, our scenario assumes 20\% client participation per communication round, a design choice that reflects the partial availability of devices in real-world Metaverse settings due to factors such as network latency, device idle states, or energy constraints. We run 100 communication rounds to adequately capture the convergence dynamics of the system under fluctuating carbon intensity levels. Furthermore, setting 5 local training epochs per client strikes a practical balance between local computation and global communication frequency, an essential trade-off for sustaining real-time responsiveness in latency-sensitive Metaverse applications.

\subsection{Ablation Insights}
This ablation enables us to answer the following research questions:

\begin{itemize}
    \item \textit{How much does RL-based orchestration contribute to convergence and accuracy under constrained and heterogeneous settings?}
    \item \textit{Can carbon-aware scheduling reduce emissions without degrading model performance?}
    \item \textit{What is the effect of combining both strategies versus deploying them individually?}
\end{itemize}

\section{Results and Discussion}

To assess the effectiveness of the proposed MetaFed framework, we conducted a systematic evaluation against established FL baselines. Experiments were carried out on the MNIST and CIFAR-10 datasets, with performance evaluated across three key dimensions: classification accuracy (\%), carbon emissions (gCO\textsubscript{2} per communication round), and time efficiency (average round duration in seconds). The results highlight MetaFed’s capability to jointly optimize predictive performance, sustainability, and computational efficiency.

\subsection{Performance on the MNIST Dataset}

\begin{table*}[t]
\centering
\caption{MNIST Performance Comparison: Classification Accuracy, CO\textsubscript{2} Emissions, and Average Round Time}
\label{tab:mnist_results}
\begin{tabular}{lcccc}
\toprule
\textbf{Model}       & \textbf{Accuracy (\%)} & \textbf{CO\textsubscript{2} (g/round)} & \textbf{Time (s/round)} & \textbf{Cumulative CO\textsubscript{2} (g)} \\
\midrule
MetaFed (RL + Green + RT)       & \textbf{99.60}         & \textbf{337.55}                       & 33.88                   & \textbf{45,846}                                    \\
MetaFed (RL + RT)         & 99.54                  & 580.40                                & 35.42                   & 57,821                                 \\
MetaFed (Green + RT)           & 99.09                  & 386.70                                & \textbf{31.82}          & \textbf{45,826}                                 \\
FedAvg (RT)               & 99.19                  & 578.40                                & 33.44                   & 57,755                                  \\
FedProx (RT)              & 99.14                  & 578.10                                & 46.85                   & 57,820                                  \\
FedAdam (RT)              & 98.87                  & 575.80                                & 30.62                   & 57,719                                  \\
\bottomrule
\end{tabular}
\end{table*}

As shown in Table~\ref{tab:mnist_results}, the full MetaFed configuration (RL + Green + RT) achieves the highest classification accuracy (99.60\%) while significantly reducing environmental impact. It lowers CO\textsubscript{2} emissions per round by approximately 41.6\% compared to standard FL baselines such as FedAvg, FedProx, and FedAdam. Over the entire training lifecycle, this translates into cumulative CO\textsubscript{2} reductions of 20.6\% compared to FedAvg (45,846g vs. 57,755g), 20.5\% compared to FedProx (45,846g vs. 57,820g), and 20.6\% compared to FedAdam (45,846g vs. 57,719g)—achieved without a notable increase in communication latency. Interestingly, MetaFed (Green + RT), which omits RL but retains carbon-aware orchestration, also yields substantial sustainability gains. It records 386.70g of CO\textsubscript{2} per round while maintaining competitive accuracy (99.09\%). Its cumulative emissions (45,826g) closely align with the full MetaFed setup, highlighting the strong standalone contribution of carbon-aware client selection strategies.

Beyond performance metrics, MetaFed also demonstrates better convergence behavior. The RL-guided orchestration mechanism supports consistent learning progress across rounds with reduced fluctuations, in contrast to traditional FL approaches. This stability highlights MetaFed’s effectiveness in achieving a balanced optimization of learning performance and environmental sustainability, making it a strong candidate for real-world Metaverse scenarios where both efficiency and ecological responsibility are critical.

\begin{figure}[htbp]
    \centering
    \includegraphics[width=0.8\linewidth]{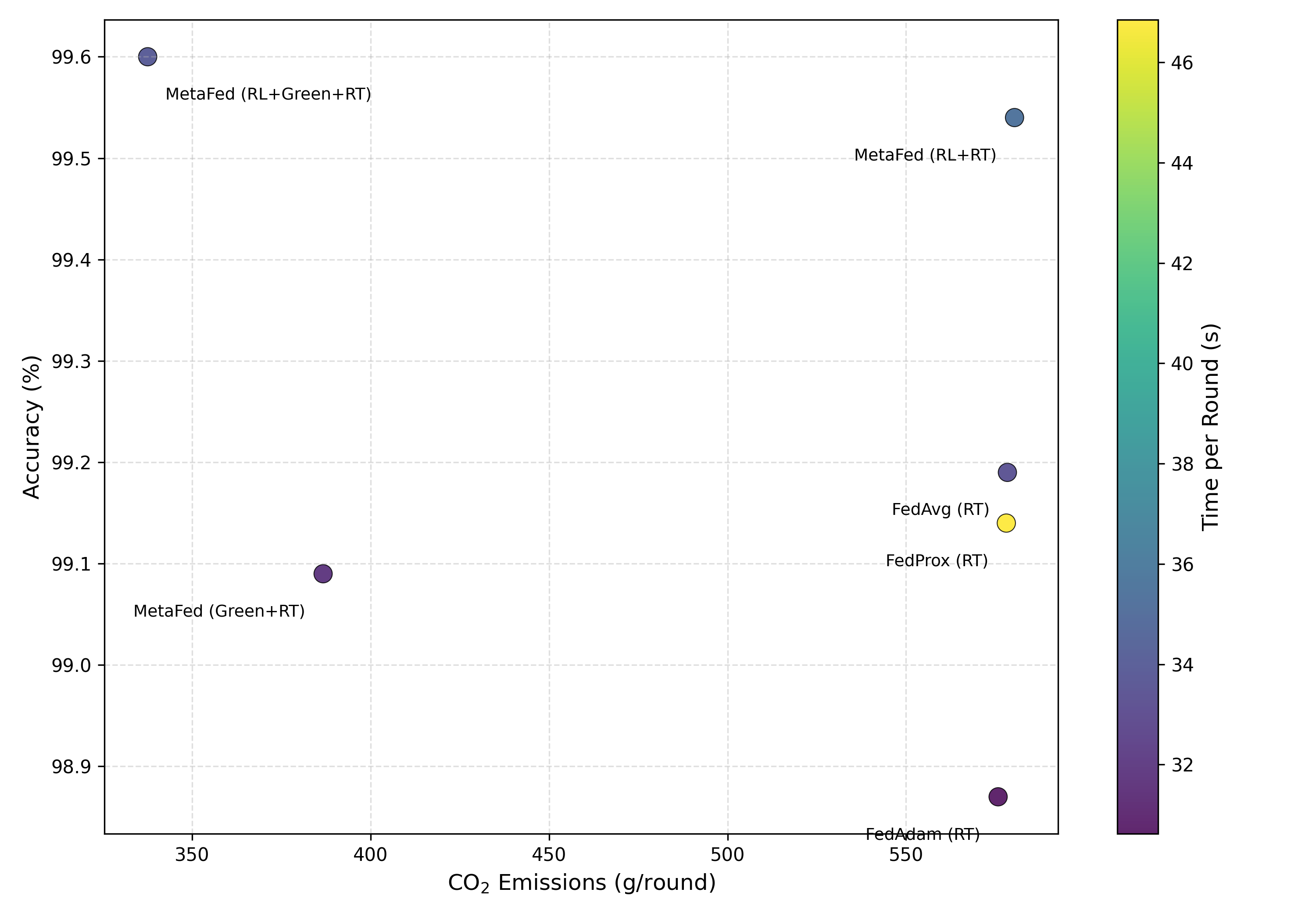}
    \caption{Accuracy–emission trade-off for MNIST: Comparison of federated learning models based on classification accuracy and per-round CO\textsubscript{2} emissions.}
    \label{fig:mnist_performance}
\end{figure}

Figure~\ref{fig:mnist_performance} illustrates the trade-off landscape between classification accuracy and carbon emissions for all evaluated methods. The clustering of MetaFed variants in the upper-left quadrant, characterized by high accuracy and low emissions, highlights the robustness and efficiency of the proposed framework. This clear separation from baseline models reinforces the effectiveness of MetaFed’s dual optimization strategy, demonstrating that enhancing environmental sustainability does not come at the cost of learning performance, even on foundational recognition tasks like MNIST.

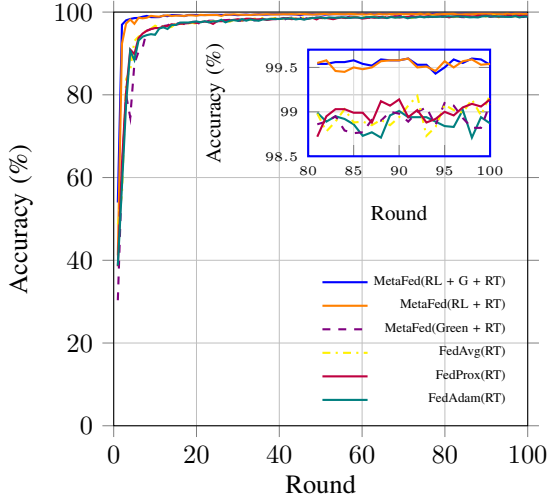
\begin{figure}[htbp]
  \centering
  \input{zoom} 
  \caption{Accuracy over communication rounds on the MNIST dataset for MetaFed and baseline federated learning models.}
  \label{fig:mnist_accuracy_zoom}
\end{figure}

The detailed performance analysis highlights the distinct contributions of each MetaFed component, as provided in Figure~\ref{fig:mnist_accuracy_zoom}. While the full MetaFed configuration delivers the most optimal results, the strong standalone performance of MetaFed (Green + RT) emphasizes the effectiveness of carbon-aware optimization. These findings suggest that the framework’s advantages are not solely reliant on the complexity of RL-based orchestration, but are rooted in fundamental enhancements to resource-conscious FL design.

\vspace{0.5em}

\subsection{Performance on the CIFAR-10 Dataset}

On the more complex CIFAR-10 dataset, the full MetaFed configuration (RL + Green + RT) achieves the highest classification accuracy at 80.26\%, marking a relative improvement of 20.5\% over FedAvg and 17.0\% over FedProx. As detailed in Table~\ref{tab:cifar_results}, this performance gain is accompanied by substantial improvements in environmental efficiency. MetaFed reduces per-round CO\textsubscript{2} emissions by approximately 49.9\% (287.90g vs. 575.80g) compared to these baselines. Cumulatively, it achieves a 21.0\% reduction in total emissions compared to FedAvg (45,634g vs. 57,755g) and 21.1\% compared to FedProx (45,634g vs. 57,820g). While MetaFed incurs a modest increase in average round time, approximately 3.7 seconds longer than FedAvg, the significant gains in both sustainability and model accuracy clearly justify this minor trade-off. The synergistic integration of RL and carbon-aware orchestration ensures optimal client selection that not only maximizes learning outcomes but also minimizes environmental footprint throughout the federated training process.

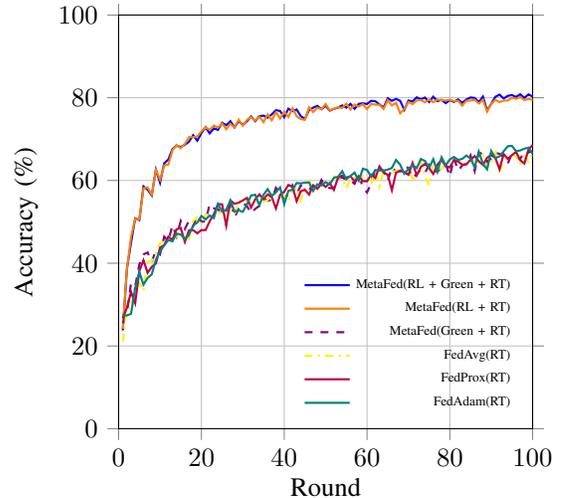
\begin{figure}[htbp]
  \centering
  \input{cifaracc} 
  \caption{Accuracy over communication rounds on the CIFAR-10 dataset for MetaFed and baseline federated learning models.}
  \label{fig:cifar_accuracy}
\end{figure}

The convergence behavior on CIFAR-10 further highlights the effectiveness of MetaFed. The RL module enables dynamic adaptation in client orchestration, facilitating faster convergence and earlier stabilization than traditional FL approaches. The round-wise accuracy progression exhibits sharper initial improvements and more consistent plateauing, indicating enhanced learning efficiency and robustness under heterogeneous data conditions, as shown in Figure~\ref{fig:cifar_accuracy}.

\vspace{0.35em}

\begin{table*}[t]
\centering
\caption{CIFAR-10 Performance Comparison: Classification Accuracy, CO\textsubscript{2} Emissions, and Average Round Time}
\label{tab:cifar_results}
\begin{tabular}{lcccc}
\toprule
\textbf{Model}       & \textbf{Accuracy (\%)} & \textbf{CO\textsubscript{2} (g/round)} & \textbf{Time (s/round)} & \textbf{Cumulative CO\textsubscript{2} (g)} \\
\midrule
MetaFed (RL + Green + RT)        & \textbf{80.26}         & \textbf{287.90}                       & 30.30                   & \textbf{45,634}                                    \\
MetaFed (RL + RT)          & 79.46                  & 575.80                                & 28.25                   & 57,545                                    \\
MetaFed (Green + RT)           & 67.12                  & \textbf{287.90}                       & \textbf{27.16}          & \textbf{45,826}                                  \\
FedAvg (RT)             & 66.56                  & 575.70                                & 26.60                   & 57,755                                  \\
FedProx (RT)               & 68.60                  & 575.80                                & 37.88                   & 57,820                                  \\
FedAdam (RT)               & 66.40                  & 575.70                                & \textbf{27.39}                   & 57,719                                  \\
\bottomrule
\end{tabular}
\end{table*}

\begin{figure}[htbp]
    \centering
    \includegraphics[width=0.8\linewidth]{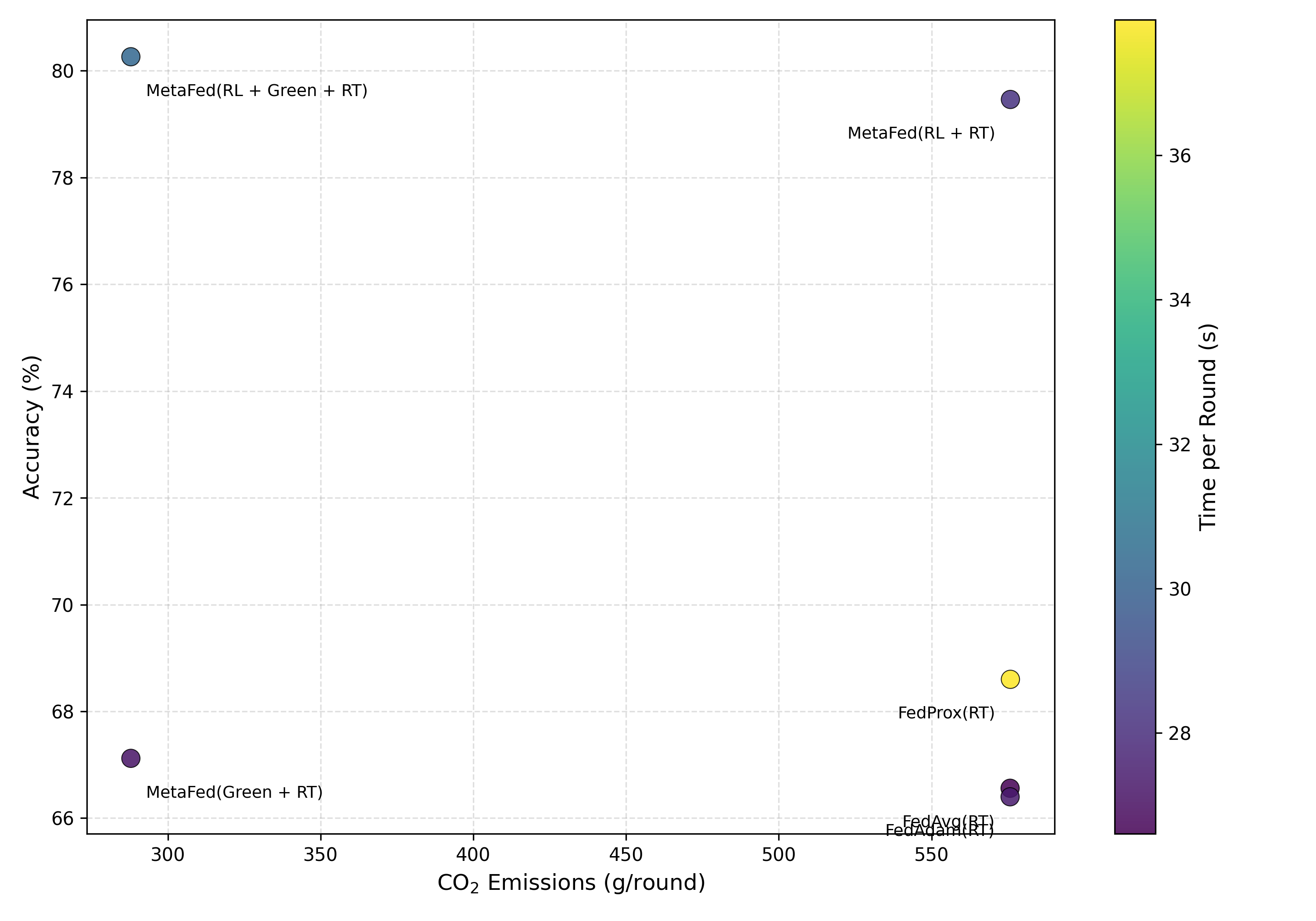}
    \caption{Accuracy–emission trade-off for  CIFAR-10: Comparison of federated learning models based on classification accuracy and per-round CO\textsubscript{2} emissions.}
    \label{fig:cifar_performance}
\end{figure}

Figure~\ref{fig:cifar_performance} illustrates the trade-off landscape between model performance and carbon emissions. MetaFed variants are distinctly positioned in the upper-left quadrant, representing high accuracy and low environmental cost, clearly separating them from baseline methods. This visualization affirms MetaFed’s capability to overcome the conventional trade-off between predictive accuracy and sustainability, setting a new standard for environmentally responsible FL in complex visual recognition scenarios.

\subsection{Multi-Objective Optimization Analysis}

The evaluation results demonstrate MetaFed (RL + Green + RT)’s effectiveness in balancing the competing objectives of model accuracy, environmental sustainability, and training efficiency. Key observations include:

\begin{itemize}
\item \textbf{Predictive Performance:} The integration of MARL enhances client orchestration, resulting in improved convergence rates and generalization. This is evident in the accuracy margins across both datasets.

\item \textbf{Sustainability:} MetaFed’s carbon-aware scheduling substantially lowers CO\textsubscript{2} emissions without altering the model architecture or increasing the number of communication rounds, enabling greener training by design.

\item \textbf{Time Efficiency:} Despite incorporating sustainability constraints, MetaFed maintains competitive communication times. Remarkably, while FedAvg achieves the shortest average round duration on CIFAR-10 (26.60 seconds), MetaFed (Green + RT) closely follows with a comparable time of 27.16 seconds, while simultaneously reducing emissions by nearly 50\%.
\end{itemize}

\subsection{Ablation Study and Design Insights}
The ablation analysis confirms that both the MARL component and the carbon-aware scheduling heuristics independently enhance MetaFed’s performance. The comparison between MetaFed (RL + RT) and MetaFed (RL + Green + RT) highlights the added value of incorporating environmental feedback into RL-driven orchestration. This integration delivers notable gains across multiple dimensions:

\begin{itemize}
\item Up to \textbf{0.73\%} improvement in classification accuracy relative to FedAdam,
\item Up to \textbf{49.9\%} reduction in per-round CO\textsubscript{2} emissions compared to FedAvg,
\item Comparable or faster average communication times, with differences within \textbf{3.2 seconds}.
\end{itemize}

These findings emphasize the synergistic impact of combining sustainability-aware signals with intelligent orchestration policies, offering a scalable path toward high-performing and environmentally responsible FL.

\subsection{Implications and Future Directions}

The empirical findings establish MetaFed (RL + Green + RT) as a promising pathway for deploying FL in carbon-regulated environments, including sectors such as healthcare, finance, and edge-powered applications within the Metaverse. By uniting MARL with sustainability-aware orchestration, MetaFed emerges as a scalable and environmentally responsible alternative to conventional FL approaches.

\subsubsection{MetaFed’s Role in Enabling Rich Metaverse Experiences}

MetaFed’s orchestration capabilities have direct implications for enhancing the responsiveness, security, and environmental sustainability of next-generation Metaverse applications:

\begin{itemize}
\item \textbf{Ultra-Low Latency XR Support:} MetaFed enables sub-20ms responsiveness by adaptively allocating compute resources based on real-time system feedback. This ensures smooth XR interactions, prevents motion sickness, and maintains immersion in multi-user, high-complexity environments.

\item \textbf{Privacy-Preserving Personalization:} By leveraging differential privacy and homomorphic encryption, MetaFed facilitates secure, on-device processing of sensitive biometric and behavioral data. This supports personalized avatar generation, adaptive interfaces, and context-aware virtual environments, all while maintaining user privacy and regulatory compliance.

\item \textbf{Sustainable Large-Scale Simulations:} The integration of carbon-aware scheduling allows persistent, large-scale virtual environments to operate with minimized environmental impact. This capability is essential for realizing environmentally sustainable Metaverse platforms hosting thousands of concurrent users.

\item \textbf{Adaptive Content Quality:} MetaFed’s MARL engine can dynamically tune rendering fidelity, simulation complexity, and AI-driven content generation based on real-time green energy availability. This enables high-quality user experiences while aligning with sustainability goals.
\end{itemize}

\subsubsection{Implications for Broader Metaverse Infrastructure}

\begin{figure}[t]
    \centering
    \includegraphics[width=0.97\linewidth]{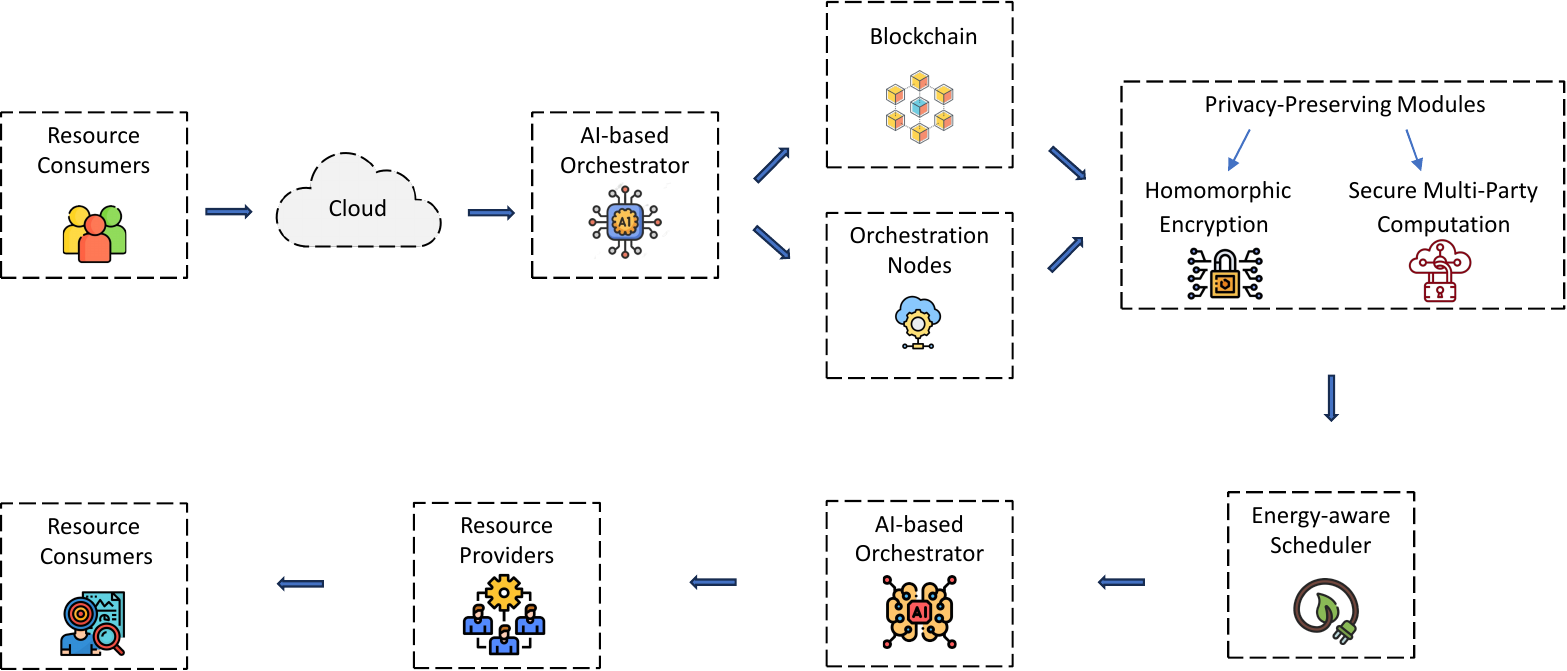}
    \caption{MetaFed architecture showing interaction between resource providers, federated agents, and carbon-aware scheduling.}
    \label{fig:system_architecture}
\end{figure}

MetaFed’s multi-objective optimization paradigm offers a blueprint for sustainable infrastructure across the Metaverse ecosystem, as presented in Figure~\ref{fig:system_architecture}. The core design principles of balancing performance, privacy, and sustainability can serve as a foundation for developing other Metaverse subsystems, such as decentralized storage, content delivery networks, and real-time communication frameworks.

\subsubsection{Future Research Directions}

Building on the current foundation, future research will explore several promising avenues:

\begin{itemize}
\item \textbf{Energy-Aware Scheduling with Market Dynamics:} Incorporating real-time energy pricing signals into scheduling decisions to optimize costs and emissions simultaneously.

\item \textbf{Federated Trust Calibration:} Designing mechanisms for assessing and adjusting trust in heterogeneous and unreliable client environments.

\item \textbf{Blockchain-Based Incentivization:} Integrating smart contracts and decentralized tokens to reward sustainable resource contributions in federated systems.

\item \textbf{Support for Emerging Modalities:} Extending MetaFed to accommodate new interfaces in the Metaverse, such as haptic feedback and neural input systems, while ensuring sustainability and privacy are preserved.
\end{itemize}

\section{Conclusion}

This work presented \textit{MetaFed}, a FL framework that integrates RL and carbon-aware orchestration to jointly optimize performance, sustainability, and efficiency. Experimental results on MNIST and CIFAR-10 confirm its effectiveness in reducing emissions while maintaining high accuracy and low latency. MetaFed’s design offers a scalable and environmentally responsible approach for real-world applications, including healthcare and the Metaverse. Its principles also provide a foundation for future research into sustainable infrastructure, privacy-preserving personalization, and adaptive system design in distributed environments.

\bibliographystyle{IEEEtran}

\end{document}

%% file: zoom.tex

\begin{tikzpicture}
  \begin{axis}[
    name=main,
    width=0.80\columnwidth,
    height=0.80\columnwidth,
    xlabel={Round},
    ylabel={Accuracy (\%)},
    xmin=0, xmax=100,
    ymin=0, ymax=100,
    grid=both,
    grid style={line width=.1pt, draw=gray!20},
    major grid style={line width=.2pt, draw=gray!50},
      legend style={
      at={(0.98,0.02)},
      anchor=south east,
      font=\tiny,
      draw=none,
      fill=none
    },
    legend cell align=right,
    tick align=outside
  ]

    \addplot[thick, color=blue, mark=none] coordinates {
      (1,53.97)  (2,96.96)  (3,98.01)  (4,98.33)  (5,98.55)
      (6,98.59)  (7,98.88)  (8,99.07)  (9,98.78) (10,98.77)
      (11,98.79) (12,99.14) (13,99.01) (14,99.27) (15,99.28)
      (16,99.20) (17,99.03) (18,99.29) (19,99.31) (20,99.29)
      (21,99.28) (22,99.29) (23,99.32) (24,99.34) (25,99.34)
      (26,99.32) (27,99.31) (28,99.37) (29,99.33) (30,99.25)
      (31,99.34) (32,99.41) (33,99.32) (34,99.38) (35,99.47)
      (36,99.39) (37,99.39) (38,99.47) (39,99.48) (40,99.47)
      (41,99.49) (42,99.48) (43,99.45) (44,99.42) (45,99.27)
      (46,99.47) (47,99.48) (48,99.49) (49,99.44) (50,99.50)
      (51,99.38) (52,99.44) (53,99.49) (54,99.51) (55,99.48)
      (56,99.47) (57,99.53) (58,99.46) (59,99.55) (60,99.50)
      (61,99.55) (62,99.56) (63,99.46) (64,99.52) (65,99.57)
      (66,99.50) (67,99.60) (68,99.46) (69,99.55) (70,99.60)
      (71,99.57) (72,99.56) (73,99.62) (74,99.52) (75,99.55)
      (76,99.50) (77,99.57) (78,99.62) (79,99.51) (80,99.58)
      (81,99.54) (82,99.54) (83,99.56) (84,99.56) (85,99.58)
      (86,99.54) (87,99.52) (88,99.59) (89,99.58) (90,99.58)
      (91,99.60) (92,99.53) (93,99.53) (94,99.43) (95,99.50)
      (96,99.59) (97,99.55) (98,99.60) (99,99.59) (100,99.53)
    };
    \addlegendentry{MetaFed(RL + G + RT)}
  
    \addplot[thick, color=orange, mark=none] coordinates {
      (1,41.18)  (2,92.37)  (3,97.25)  (4,98.20)  (5,97.17)
      (6,98.62)  (7,98.49)  (8,98.64)  (9,99.08) (10,98.96)
      (11,98.89) (12,99.11) (13,99.23) (14,99.05) (15,99.15)
      (16,99.13) (17,99.07) (18,99.24) (19,99.21) (20,99.12)
      (21,99.32) (22,99.25) (23,99.28) (24,99.28) (25,99.29)
      (26,99.07) (27,99.40) (28,99.35) (29,99.46) (30,99.38)
      (31,99.35) (32,99.36) (33,99.30) (34,99.42) (35,99.41)
      (36,99.42) (37,99.49) (38,99.43) (39,99.43) (40,99.45)
      (41,99.49) (42,99.47) (43,99.52) (44,99.46) (45,99.47)
      (46,99.54) (47,99.47) (48,99.41) (49,99.46) (50,99.49)
      (51,99.54) (52,99.44) (53,99.52) (54,99.52) (55,99.49)
      (56,99.48) (57,99.54) (58,99.51) (59,99.46) (60,99.49)
      (61,99.42) (62,99.52) (63,99.38) (64,99.52) (65,99.48)
      (66,99.46) (67,99.55) (68,99.46) (69,99.53) (70,99.50)
      (71,99.49) (72,99.52) (73,99.50) (74,99.57) (75,99.44)
      (76,99.55) (77,99.36) (78,99.51) (79,99.48) (80,99.55)
      (81,99.55) (82,99.58) (83,99.46) (84,99.45) (85,99.50)
      (86,99.48) (87,99.50) (88,99.57) (89,99.58) (90,99.58)
      (91,99.60) (92,99.50) (93,99.51) (94,99.46) (95,99.57)
      (96,99.50) (97,99.57) (98,99.59) (99,99.53) (100,99.54)
    };
    \addlegendentry{MetaFed(RL + RT)}

    \addplot[thick,color=violet, dashed, mark=none] coordinates {
      (1,30.34) (2,61.30) (3,79.90) (4,74.55) (5,87.57)
      (6,90.39) (7,91.41) (8,95.49) (9,95.89) (10,96.23)
      (11,96.48) (12,96.62) (13,96.79) (14,97.29) (15,97.16)
      (16,97.35) (17,97.11) (18,97.31) (19,97.64) (20,97.65)
      (21,97.01) (22,97.45) (23,97.91) (24,97.91) (25,97.75)
      (26,98.06) (27,98.05) (28,98.12) (29,97.73) (30,97.89)
      (31,97.89) (32,98.18) (33,98.18) (34,97.99) (35,98.15)
      (36,98.20) (37,98.31) (38,98.36) (39,98.28) (40,98.35)
      (41,98.06) (42,98.53) (43,98.65) (44,98.49) (45,98.65)
      (46,98.31) (47,98.61) (48,98.74) (49,98.69) (50,98.56)
      (51,98.56) (52,98.46) (53,98.40) (54,98.67) (55,98.57)
      (56,98.60) (57,98.90) (58,98.78) (59,98.60) (60,98.58)
      (61,98.71) (62,98.61) (63,98.61) (64,98.77) (65,98.87)
      (66,98.69) (67,98.82) (68,98.88) (69,99.03) (70,98.94)
      (71,98.82) (72,98.98) (73,98.79) (74,99.00) (75,98.87)
      (76,98.79) (77,98.92) (78,98.95) (79,98.96) (80,98.83)
      (81,98.86) (82,98.89) (83,98.95) (84,98.79) (85,98.76)
      (86,98.77) (87,98.91) (88,98.90) (89,99.00) (90,98.98)
      (91,98.89) (92,99.00) (93,99.05) (94,98.85) (95,99.10)
      (96,99.07) (97,98.93) (98,98.82) (99,98.82) (100,99.09)
    }; 
    \addlegendentry{MetaFed(Green + RT)}

    \addplot[thick, color=yellow, dash dot, mark=none] coordinates {
      (1,48.66) (2,56.08) (3,83.77) (4,86.03) (5,93.02)
      (6,94.08) (7,94.45) (8,95.86) (9,95.98) (10,96.71)
      (11,96.20) (12,96.52) (13,96.21) (14,96.40) (15,97.23)
      (16,97.21) (17,97.04) (18,97.27) (19,97.39) (20,97.18)
      (21,97.52) (22,97.82) (23,97.61) (24,97.86) (25,97.76)
      (26,97.96) (27,98.03) (28,97.66) (29,98.10) (30,97.92)
      (31,97.97) (32,98.22) (33,98.13) (34,98.42) (35,98.31)
      (36,98.57) (37,98.29) (38,98.37) (39,98.58) (40,98.44)
      (41,98.31) (42,98.57) (43,98.46) (44,98.42) (45,98.56)
      (46,98.66) (47,98.61) (48,98.55) (49,98.59) (50,98.29)
      (51,98.68) (52,98.59) (53,98.45) (54,98.69) (55,98.55)
      (56,98.67) (57,98.75) (58,98.88) (59,98.80) (60,98.93)
      (61,98.72) (62,98.89) (63,98.78) (64,98.81) (65,98.96)
      (66,98.49) (67,98.91) (68,98.73) (69,98.77) (70,98.86)
      (71,98.87) (72,98.56) (73,98.87) (74,98.83) (75,98.79)
      (76,99.00) (77,99.02) (78,98.96) (79,99.01) (80,98.92)
      (81,98.98) (82,98.78) (83,98.89) (84,99.03) (85,98.88)
      (86,98.89) (87,98.85) (88,98.91) (89,98.85) (90,98.94)
      (91,99.07) (92,99.18) (93,98.73) (94,98.83) (95,99.08)
      (96,99.02) (97,98.99) (98,99.14) (99,98.96) (100,99.19)
    }; 
    \addlegendentry{FedAvg(RT)}

    \addplot[thick,color=purple, solid, mark=none] coordinates {
      (1,38.55) (2,63.50) (3,80.75) (4,90.92) (5,89.86)
      (6,93.70) (7,94.82) (8,95.48) (9,95.90) (10,96.14)
      (11,96.14) (12,96.49) (13,96.62) (14,97.06) (15,96.93)
      (16,97.17) (17,97.23) (18,97.32) (19,97.57) (20,97.37)
      (21,97.59) (22,97.70) (23,97.60) (24,97.14) (25,97.81)
      (26,97.79) (27,97.85) (28,98.00) (29,98.12) (30,98.28)
      (31,98.13) (32,98.39) (33,98.38) (34,98.25) (35,98.22)
      (36,98.35) (37,98.09) (38,98.23) (39,98.37) (40,98.31)
      (41,98.57) (42,98.46) (43,98.29) (44,98.41) (45,98.34)
      (46,98.32) (47,98.38) (48,98.62) (49,98.23) (50,98.72)
      (51,98.61) (52,98.47) (53,98.28) (54,98.54) (55,98.68)
      (56,98.56) (57,98.67) (58,98.71) (59,98.70) (60,98.42)
      (61,98.54) (62,98.74) (63,98.62) (64,98.76) (65,98.65)
      (66,98.66) (67,98.89) (68,98.94) (69,98.69) (70,98.80)
      (71,98.83) (72,98.92) (73,98.84) (74,98.96) (75,98.84)
      (76,99.01) (77,98.81) (78,98.94) (79,98.97) (80,98.99)
      (81,98.72) (82,98.95) (83,99.03) (84,99.03) (85,98.99)
      (86,98.99) (87,98.88) (88,99.12) (89,99.07) (90,99.14)
      (91,98.94) (92,99.02) (93,98.88) (94,98.92) (95,99.00)
      (96,98.96) (97,99.05) (98,99.09) (99,99.06) (100,99.14)
    };
    \addlegendentry{FedProx(RT)}

    \addplot[thick, color=teal, solid, mark=none] coordinates {
      (1,38.72) (2,58.04) (3,77.83) (4,90.62) (4,90.62) (5,88.40)
      (6,93.12) (7,94.07) (8,94.43) (9,94.66) (10,94.51)
      (11,96.02) (12,96.32) (13,95.91) (14,96.87) (15,96.89)
      (16,97.46) (17,97.48) (18,97.47) (19,97.59) (20,97.73)
      (21,97.54) (22,97.79) (23,97.51) (24,97.78) (25,97.73)
      (26,98.14) (27,97.52) (28,97.90) (29,98.07) (30,97.88)
      (31,98.30) (32,97.92) (33,98.22) (34,98.03) (35,97.90)
      (36,98.22) (37,98.22) (38,98.33) (39,98.33) (40,98.34)
      (41,98.45) (42,98.42) (43,98.48) (44,98.23) (45,98.39)
      (46,98.17) (47,98.43) (48,98.56) (49,98.61) (50,98.76)
      (51,98.50) (52,98.71) (53,98.64) (54,98.61) (55,98.75)
      (56,98.69) (57,98.59) (58,98.72) (59,98.68) (60,98.56)
      (61,98.76) (62,98.78) (63,98.77) (64,98.60) (65,98.77)
      (66,98.55) (67,98.81) (68,98.65) (69,98.76) (70,98.82)
      (71,98.50) (72,98.72) (73,98.70) (74,98.89) (75,98.73)
      (76,98.87) (77,98.87) (78,98.84) (79,98.90) (80,98.90)
      (81,98.99) (82,98.89) (83,98.95) (84,98.92) (85,98.86)
      (86,98.73) (87,98.77) (88,98.71) (89,98.96) (90,99.01)
      (91,98.94) (92,98.94) (93,98.94) (94,98.89) (95,98.84)
      (96,98.83) (97,99.02) (98,98.71) (99,98.94) (100,98.87)
    }; 
    \addlegendentry{FedAdam(RT)}

  \end{axis}
  
  \begin{axis}[
    at={(main.north east)},
    anchor=north east,
    xshift=-0.5cm,
    yshift=-0.5cm,
    width=4cm,
    height=3cm,
    xlabel={\footnotesize Round},
    ylabel={\footnotesize Accuracy (\%)},
    xmin=80, xmax=100,
    ymin=98.5, ymax=99.7,
    grid=both,
    grid style={line width=.1pt, draw=gray!30},
    tick label style={font=\tiny},
    label style={font=\tiny},
    title={\small },
    title style={font=\small},
    draw=blue,
     thick
  ]
    
    \addplot[thick, color=teal, solid, mark=none] coordinates {
      (81,98.99) (82,98.89) (83,98.95) (84,98.92) (85,98.86)
      (86,98.73) (87,98.77) (88,98.71) (89,98.96) (90,99.01)
      (91,98.94) (92,98.94) (93,98.94) (94,98.89) (95,98.84)
      (96,98.83) (97,99.02) (98,98.71) (99,98.94) (100,98.87)
    }; 

    \addplot[thick,color=violet, dashed, mark=none] coordinates {
      (81,98.86) (82,98.89) (83,98.95) (84,98.79) (85,98.76)
      (86,98.77) (87,98.91) (88,98.90) (89,99.00) (90,98.98)
      (91,98.89) (92,99.00) (93,99.05) (94,98.85) (95,99.10)
      (96,99.07) (97,98.93) (98,98.82) (99,98.82) (100,99.09)
    }; 

    \addplot[thick, color=yellow, dash dot, mark=none] coordinates {
      (81,98.98) (82,98.78) (83,98.89) (84,99.03) (85,98.88)
      (86,98.89) (87,98.85) (88,98.91) (89,98.85) (90,98.94)
      (91,99.07) (92,99.18) (93,98.73) (94,98.83) (95,99.08)
      (96,99.02) (97,98.99) (98,99.14) (99,98.96) (100,99.19)
    }; 

    \addplot[thick,color=purple, solid, mark=none] coordinates {
      (81,98.72) (82,98.95) (83,99.03) (84,99.03) (85,98.99)
      (86,98.99) (87,98.88) (88,99.12) (89,99.07) (90,99.14)
      (91,98.94) (92,99.02) (93,98.88) (94,98.92) (95,99.00)
      (96,98.96) (97,99.05) (98,99.09) (99,99.06) (100,99.14)
    };

    \addplot[thick, color=blue, mark=none] coordinates {
      (81,99.54) (82,99.54) (83,99.56) (84,99.56) (85,99.58)
      (86,99.54) (87,99.52) (88,99.59) (89,99.58) (90,99.58)
      (91,99.60) (92,99.53) (93,99.53) (94,99.43) (95,99.50)
      (96,99.59) (97,99.55) (98,99.60) (99,99.59) (100,99.53)
    };

    \addplot[thick, color=orange, mark=none] coordinates {
      (81,99.55) (82,99.58) (83,99.46) (84,99.45) (85,99.50)
      (86,99.48) (87,99.50) (88,99.57) (89,99.58) (90,99.58)
      (91,99.60) (92,99.50) (93,99.51) (94,99.46) (95,99.57)
      (96,99.50) (97,99.57) (98,99.59) (99,99.53) (100,99.54)
    };
    
\end{axis}
  
\end{tikzpicture}

%% file: cifaracc.tex
\begin{tikzpicture}
  \begin{axis}[
       width=0.80\columnwidth,
    height=0.80\columnwidth,
    xlabel={Round},
    ylabel={Accuracy (\%)},
    xmin=0, xmax=100,
    ymin=0, ymax=100,
    grid=both,
    grid style={line width=.1pt, draw=gray!20},
    major grid style={line width=.2pt, draw=gray!50},
    legend style={
      at={(0.98,0.02)},
      anchor=south east,
      font=\tiny,
      draw=none,
      fill=none
    },
    legend cell align=right,
    tick align=outside,
    spy using outlines={rectangle, magnification=3, size=3cm, connect spies}
  ]

    \addplot[thick, color=blue, solid, mark=none] coordinates {
      (1,23.75)   (2,38.65)   (3,44.97)   (4,50.85)   (5,50.42)
      (6,58.57)   (7,57.79)   (8,56.38)   (9,62.62)   (10,59.47)
      (11,64.16)  (12,65.05)  (13,67.59)  (14,68.39)  (15,68.10)
      (16,68.63)  (17,69.39)  (18,70.93)  (19,69.62)  (20,71.46)
      (21,72.87)  (22,71.37)  (23,72.59)  (24,72.23)  (25,73.96)
      (26,73.36)  (27,74.32)  (28,72.82)  (29,74.33)  (30,73.46)
      (31,74.31)  (32,75.49)  (33,74.33)  (34,75.56)  (35,75.67)
      (36,75.17)  (37,76.00)  (38,77.00)  (39,76.80)  (40,75.27)
      (41,77.36)  (42,76.74)  (43,77.15)  (44,75.54)  (45,74.89)
      (46,77.04)  (47,77.60)  (48,78.04)  (49,77.22)  (50,78.06)
      (51,76.88)  (52,77.32)  (53,77.61)  (54,78.14)  (55,77.61)
      (56,78.52)  (57,77.32)  (58,78.52)  (59,78.65)  (60,78.49)
      (61,79.25)  (62,79.16)  (63,78.38)  (64,77.62)  (65,79.88)
      (66,79.25)  (67,79.72)  (68,79.27)  (69,77.09)  (70,79.33)
      (71,79.12)  (72,80.10)  (73,79.64)  (74,80.04)  (75,78.96)
      (76,80.25)  (77,79.18)  (78,79.49)  (79,78.68)  (80,79.44)
      (81,78.98)  (82,79.61)  (83,78.73)  (84,79.59)  (85,80.06)
      (86,79.95)  (87,78.93)  (88,80.28)  (89,76.92)  (90,78.51)
      (91,80.22)  (92,80.75)  (93,79.82)  (94,80.45)  (95,80.65)
      (96,80.17)  (97,80.85)  (98,79.95)  (99,80.81)  (100,80.26)
    };
    \addlegendentry{MetaFed(RL + Green + RT)}

    \addplot[thick, color=orange, solid, mark=none] coordinates {
      (1,24.15)   (2,38.54)   (3,46.40)   (4,50.76)   (5,50.57)
      (6,58.14)   (7,58.32)   (8,56.55)   (9,62.58)   (10,60.21)
      (11,63.65)  (12,63.83)  (13,67.33)  (14,68.60)  (15,67.81)
      (16,68.80)  (17,70.48)  (18,70.48)  (19,70.64)  (20,71.86)
      (21,72.87)  (22,72.23)  (23,73.29)  (24,72.77)  (25,73.78)
      (26,72.39)  (27,73.72)  (28,72.78)  (29,74.69)  (30,73.22)
      (31,74.67)  (32,75.32)  (33,74.47)  (34,75.42)  (35,74.12)
      (36,75.97)  (37,74.55)  (38,76.50)  (39,74.96)  (40,76.65)
      (41,75.37)  (42,77.45)  (43,75.53)  (44,74.82)  (45,74.65)
      (46,77.68)  (47,76.55)  (48,77.60)  (49,77.22)  (50,78.47)
      (51,76.93)  (52,77.36)  (53,77.71)  (54,77.51)  (55,78.58)
      (56,77.15)  (57,78.35)  (58,77.86)  (59,78.44)  (60,77.26)
      (61,78.51)  (62,78.34)  (63,78.30)  (64,77.27)  (65,79.42)
      (66,78.15)  (67,78.78)  (68,76.24)  (69,77.30)  (70,78.97)
      (71,78.47)  (72,79.33)  (73,79.44)  (74,78.90)  (75,79.57)
      (76,78.91)  (77,79.54)  (78,79.40)  (79,79.36)  (80,79.51)
      (81,79.18)  (82,79.25)  (83,78.08)  (84,79.27)  (85,79.31)
      (86,79.51)  (87,78.60)  (88,79.49)  (89,76.65)  (90,78.46)
      (91,79.52)  (92,78.92)  (93,79.31)  (94,79.31)  (95,80.01)
      (96,79.75)  (97,80.15)  (98,79.45)  (99,79.70)  (100,79.46)
    };
    \addlegendentry{MetaFed(RL + RT)}

    \addplot[thick, color=violet, dashed, mark=none] coordinates {
      (1,25.52)   (2,27.95)   (3,35.25)   (4,34.45)   (5,39.71)
      (6,42.12)   (7,42.60)   (8,40.26)   (9,43.19)   (10,42.16)
      (11,46.42)  (12,45.38)  (13,49.16)  (14,46.48)  (15,50.23)
      (16,48.28)  (17,50.32)  (18,48.96)  (19,51.58)  (20,50.00)
      (21,50.12)  (22,54.32)  (23,53.28)  (24,53.03)  (25,51.89)
      (26,52.57)  (27,55.19)  (28,51.10)  (29,53.18)  (30,54.53)
      (31,53.84)  (32,52.02)  (33,53.14)  (34,53.88)  (35,55.63)
      (36,55.12)  (37,55.34)  (38,58.23)  (39,57.05)  (40,55.76)
      (41,56.62)  (42,58.81)  (43,56.62)  (44,58.82)  (45,58.60)
      (46,57.95)  (47,58.70)  (48,57.83)  (49,60.13)  (50,58.53)
      (51,58.16)  (52,58.11)  (53,60.38)  (54,60.76)  (55,60.10)
      (56,59.98)  (57,59.19)  (58,61.50)  (59,58.24)  (60,57.11)
      (61,59.31)  (62,62.64)  (63,62.87)  (64,62.40)  (65,59.90)
      (66,63.20)  (67,62.12)  (68,62.18)  (69,63.25)  (70,63.08)
      (71,60.58)  (72,61.56)  (73,62.66)  (74,61.82)  (75,63.79)
      (76,62.42)  (77,62.15)  (78,63.11)  (79,64.04)  (80,60.96)
      (81,65.45)  (82,62.37)  (83,63.64)  (84,64.04)  (85,66.07)
      (86,62.32)  (87,66.69)  (88,66.46)  (89,64.46)  (90,65.35)
      (91,66.63)  (92,66.46)  (93,65.63)  (94,64.14)  (95,66.42)
      (96,65.66)  (97,65.17)  (98,67.08)  (99,66.77)  (100,67.12)
    };
    \addlegendentry{MetaFed(Green + RT)}

 \addplot[thick, color=yellow, dash dot, mark=none] coordinates {
      (1,21.06)   (2,28.78)   (3,29.18)   (4,34.76)   (5,35.31)
      (6,33.26)   (7,40.38)   (8,40.52)   (9,42.26)   (10,44.85)
      (11,45.35)  (12,46.40)  (13,47.21)  (14,46.40)  (15,48.01)
      (16,46.45)  (17,45.97)  (18,50.18)  (19,51.14)  (20,49.43)
      (21,51.77)  (22,51.63)  (23,51.11)  (24,51.56)  (25,53.14)
      (26,51.00)  (27,54.27)  (28,51.84)  (29,54.68)  (30,55.95)
      (31,55.18)  (32,54.54)  (33,54.70)  (34,54.88)  (35,54.89)
      (36,55.26)  (37,56.39)  (38,56.90)  (39,56.03)  (40,56.30)
      (41,54.97)  (42,57.62)  (43,58.54)  (44,55.26)  (45,58.38)
      (46,59.89)  (47,57.37)  (48,58.95)  (49,58.63)  (50,59.28)
      (51,59.65)  (52,61.01)  (53,60.58)  (54,60.39)  (55,58.07)
      (56,60.45)  (57,60.65)  (58,59.62)  (59,61.12)  (60,60.79)
      (61,59.01)  (62,62.87)  (63,58.57)  (64,62.46)  (65,61.37)
      (66,62.09)  (67,62.29)  (68,62.15)  (69,62.27)  (70,62.40)
      (71,60.44)  (72,60.84)  (73,64.34)  (74,61.24)  (75,58.77)
      (76,64.59)  (77,62.68)  (78,62.90)  (79,64.05)  (80,65.21)
      (81,64.90)  (82,64.82)  (83,64.73)  (84,64.27)  (85,64.17)
      (86,64.57)  (87,64.60)  (88,66.12)  (89,63.12)  (90,63.94)
      (91,66.71)  (92,62.35)  (93,64.56)  (94,65.56)  (95,66.09)
      (96,65.92)  (97,64.08)  (98,65.80)  (99,64.53)  (100,66.56)
    };
    \addlegendentry{FedAvg(RT)}

    \addplot[thick, color=purple, solid, mark=none] coordinates {
      (1,26.73)   (2,29.23)   (3,33.04)   (4,30.32)   (5,38.12)
      (6,40.71)   (7,37.68)   (8,39.12)   (9,39.93)   (10,43.76)
      (11,44.38)  (12,46.39)  (13,45.66)  (14,48.21)  (15,48.52)
      (16,45.15)  (17,47.96)  (18,48.00)  (19,47.22)  (20,48.01)
      (21,47.99)  (22,50.14)  (23,51.86)  (24,54.32)  (25,54.21)
      (26,49.08)  (27,55.18)  (28,54.36)  (29,54.63)  (30,55.17)
      (31,55.79)  (32,53.48)  (33,55.45)  (34,56.65)  (35,55.47)
      (36,56.64)  (37,55.96)  (38,53.24)  (39,56.55)  (40,57.29)
      (41,58.93)  (42,58.53)  (43,54.98)  (44,57.40)  (45,57.76)
      (46,56.59)  (47,58.86)  (48,57.99)  (49,59.26)  (50,58.45)
      (51,59.37)  (52,57.72)  (53,60.70)  (54,61.16)  (55,60.70)
      (56,61.03)  (57,59.77)  (58,61.07)  (59,59.84)  (60,59.80)
      (61,61.41)  (62,60.58)  (63,63.41)  (64,61.92)  (65,61.36)
      (66,57.65)  (67,61.86)  (68,62.41)  (69,61.97)  (70,61.38)
      (71,64.25)  (72,61.22)  (73,63.26)  (74,64.01)  (75,62.99)
      (76,61.57)  (77,63.80)  (78,65.04)  (79,63.39)  (80,63.16)
      (81,64.91)  (82,64.57)  (83,63.83)  (84,63.87)  (85,65.02)
      (86,65.88)  (87,64.64)  (88,63.77)  (89,65.48)  (90,66.17)
      (91,65.44)  (92,62.73)  (93,66.40)  (94,65.23)  (95,66.12)
      (96,66.91)  (97,64.91)  (98,64.09)  (99,67.48)  (100,68.60)
    };
    \addlegendentry{FedProx(RT)}

    \addplot[thick, color=teal, solid, mark=none] coordinates {
      (1,27.42)   (2,27.37)   (3,27.77)   (4,34.03)   (5,37.82)
      (6,34.82)   (7,36.43)   (8,37.30)   (9,40.85)   (10,42.81)
      (11,44.29)  (12,45.59)  (13,45.29)  (14,47.16)  (15,46.93)
      (16,45.87)  (17,46.84)  (18,49.15)  (19,49.82)  (20,51.42)
      (21,50.56)  (22,51.54)  (23,50.92)  (24,54.29)  (25,52.47)
      (26,52.93)  (27,55.17)  (28,52.46)  (29,52.77)  (30,52.42)
      (31,54.26)  (32,56.39)  (33,56.55)  (34,54.02)  (35,56.55)
      (36,57.90)  (37,56.26)  (38,57.14)  (39,58.32)  (40,54.28)
      (41,57.50)  (42,57.77)  (43,59.40)  (44,59.30)  (45,59.48)
      (46,59.51)  (47,57.51)  (48,59.08)  (49,58.84)  (50,59.60)
      (51,59.04)  (52,60.67)  (53,60.88)  (54,62.24)  (55,60.30)
      (56,61.10)  (57,61.90)  (58,59.93)  (59,62.86)  (60,62.55)
      (61,60.93)  (62,62.52)  (63,62.42)  (64,62.38)  (65,62.62)
      (66,63.02)  (67,64.98)  (68,63.81)  (69,61.63)  (70,62.73)
      (71,63.73)  (72,62.97)  (73,65.19)  (74,64.72)  (75,63.09)
      (76,65.13)  (77,62.09)  (78,65.60)  (79,63.66)  (80,65.09)
      (81,65.69)  (82,64.99)  (83,65.22)  (84,66.03)  (85,65.27)
      (86,63.19)  (87,66.96)  (88,64.01)  (89,62.70)  (90,67.16)
      (91,65.41)  (92,65.83)  (93,67.05)  (94,68.28)  (95,68.37)
      (96,67.07)  (97,67.26)  (98,67.69)  (99,67.93)  (100,66.40)
    };
    \addlegendentry{FedAdam(RT)}

  \end{axis}
\end{tikzpicture}